\documentclass{article}
\usepackage[utf8]{inputenc}
\usepackage{url}
\usepackage{authblk}
\usepackage{geometry}
\usepackage{fancyhdr}
\usepackage{graphicx}
\graphicspath{ {./} }

\title{ Towards Automating the AI Operations Lifecycle}

\author{Matthew Arnold}
\author{Jeffrey Boston}
\author{Michael Desmond}
\author{Evelyn Duesterwald}
\author{~~~~~~Benjamin Elder}
\author{Anupama Murthi}
\author{Jiri Navr\'atil}
\author{Darrell Reimer\thanks{Authors listed in alphabetical order.}}
\affil{IBM Research AI\\ Yorktown Heights, NY, USA}

\date{}
\begin{document}
\maketitle
\thispagestyle{fancy}

\begin{abstract}
%


Today's AI deployments often require significant human involvement and skill in the operational stages of the model lifecycle, including pre-release testing, monitoring, problem diagnosis and model improvements. 
We present a set of enabling technologies that can be used to increase the level of automation in AI operations, thus lowering the human effort required.  Since a common source of human involvement is the need to assess the performance of deployed models, we focus on technologies for performance prediction and KPI analysis and show how they can be used to improve automation in the key stages of a typical AI operations pipeline.

\end{abstract}

\section{Introduction}

\begin{figure}[ht]
    \centering
    \includegraphics[width=\linewidth]{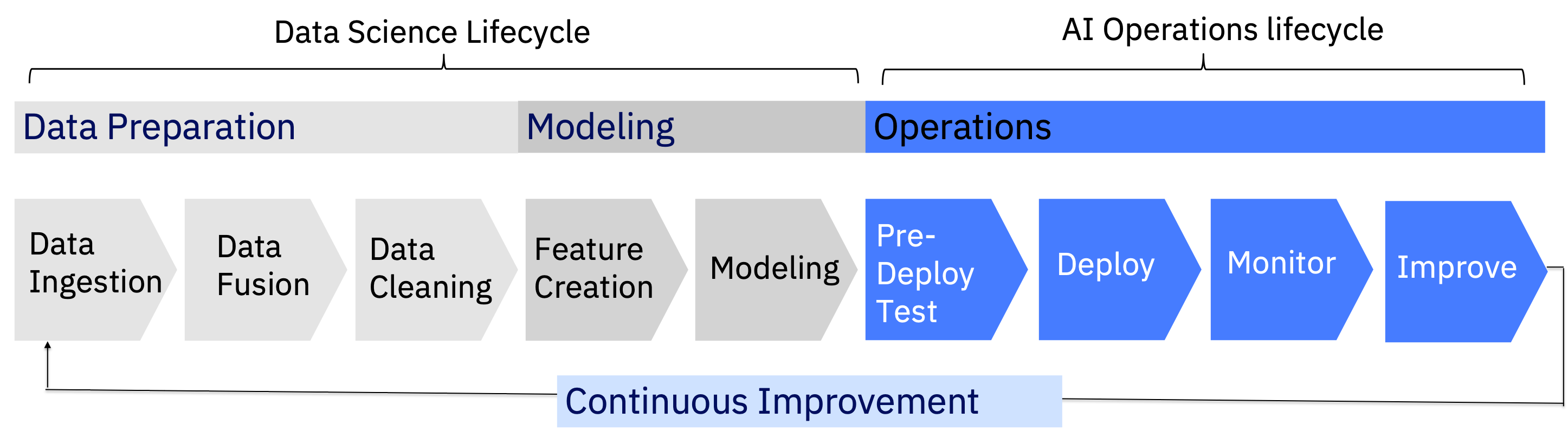}
    \caption{End-to-end AI lifecycle}
    \label{fig:e2elifecycle}
\end{figure}

The end-to-end AI lifecycle consists of many often complex stages, including data preparation, modeling, and operations.  While the details may vary from instance to instance, the overall flow often follows that depicted in Figure~\ref{fig:e2elifecycle}. 
A lot of attention in both academia and industry has been focused on the earlier data science stages of the lifecycle.
The remaining stages in AI operations are often neglected, or even overlooked entirely, despite being critical to the successful use of AI models in real-world applications.

It is not uncommon in today's AI deployments for AI operations to require significant human involvement and skill.
For example, pre-release testing generally relies on static hold-out test sets, which are expensive to create and nearly impossible to keep aligned with continually evolving production traffic~\cite{dudik2005, Zadrozny2019}.
Monitoring a model post-deployment is often achieved by performing periodic manual assessments of the log data. The cost of manual labeling during the evaluation can sometimes be reduced through crowd sourcing, but crowd sourcing may not be a viable option for enterprise customers with confidential data.  Furthermore, once a model has been identified as performing poorly, diagnosing and improving the model is a challenging, often manual, task that requires significant expertise.



We are working on addressing these operations challenges by developing a set of enabling technologies that can be used to increase the level of automation in the AI operations lifecycle, thus lowering the human effort required. A common source of human involvement across AI operations is the need to assess the performance of models deployed into production, so we focus on technologies that capture aspects of production performance:  

\begin{enumerate}
\item \textbf{Performance Prediction} - AI-based approaches for estimating accuracy-related model metrics on unlabeled data, such as the data in production traffic.
\item \textbf{KPI Analytics} - Techniques for capturing and analyzing application-level key performance indicators (KPIs) and feedback signals to monitor, analyze, and improve AI models.
\end{enumerate}

This paper describes these technologies and shows how they can be used to improve four key stages of a typical AI operations pipeline: pre-deploy test, deploy, monitor, and improve.

\section{Enabling Technologies}

This section describes enabling performance technologies developed to support the AI operations pipeline stages.

\subsection{Performance Prediction}

Performance prediction is a technique to estimate how a model will perform on a new, unlabeled data set, in terms of an accuracy-related metric (e.g, classification accuracy, regression error, or ranking precision).  Although performance prediction closely relates to the problem of uncertainty estimation (or confidence scoring), virtually all research attention in this area ~\cite{Chen2019,gal16,Geifman2017} aims at solving the task of filtering model predictions to improve its output precision.  For example, one class of such algorithms is based on calibration, i.e., producing probabilities that reflect the expected proportion of accurately classified samples \cite{GuoCalibration2017, ZadroznyCalibration2002}.  Another is a model-based approach via meta-modeling \cite{Chen2019} offering a good degree of flexibility in capturing the various sources of uncertainty.
All of these algorithms can be applied to predict a model's performance on unlabeled production (operational) data 
in a rather straightforward way: convert well-calibrated point-wise confidences generated by a model based on the above-mentioned methods into a batch accuracy prediction, thus providing insight into potential risks associated with a new version of a model, or a new operational environment. 

However, despite the clear application to production model risk management, performance prediction is rarely talked about in this context.  We believe there is significant opportunity for applying performance predictors as part of AI operations, especially for new predictors that are specifically designed for this purpose.


\subsection{KPI Analytics}

There are many metrics available for evaluating the quality of a model, such as precision, recall, F1, and accuracy.
However, these metrics evaluate a model in isolation, independent of the context of how the model is being used.
In a production setting, machine learning models are rarely exposed directly to end users;
typically, models are {\it embedded within an application}, where the application leverages the model to help make key decisions, such as whether to approve a loan or whether to tag an image.  

End users of such an application generally do not care about model performance metrics.  In fact, they may not even be aware that the application is using a model in the first place.
What ultimately matters is how well the application is performing {\it as visible to the end user}.
User-visible application performance is generally tracked with {\it business metrics}, or {\it key performance indicators} (KPI), e.g. sales rate, click rate, customer satisfaction, and time on page.

Thus, when evaluating models in the context of an application, model performance metrics are not sufficient. Instead, it is critical to understand how the model behavior is impacting the application's KPIs.
Failure to do so can result in wasting resources to improve model metrics that in fact had little or no impact on what matters to end users.
The most basic step towards supporting such KPI-based analyses is to ensure that KPIs and model metrics are being stored with a common correlation ID to identify which model operations contributed to transactions with a particular KPI score.   

\section{Enabled AI-Operations Stages}
This section describes how the above enabling technologies can be leveraged within a typical AI Operations pipeline.


\subsection{Pre-deploy Test}

The objective of the pre-deploy test stage is to assess a model's level of readiness for deployment into production~\cite{zhang2019machine}. 
This is traditionally performed with static test sets~\cite{Ishikawa2018}, which are laborious to setup, expensive to maintain, and notorious for being out of date, since data in production is continually changing over time.  Relying on static test sets to assess model quality, therefore, often leads to poor results in real world deployments.

Our AI enabled pre-deploy testing approach augments traditional test sets with performance prediction.  Instead of evaluating a model against a test set and using that test set score as an indicator of model quality, the test set is used to train a performance predictor, and the predictor is run on a recent window of production traffic. The resulting scores from the predictor provide a performance estimate of the model on actual production traffic. 

While this approach still requires some amount of labeled data, it has many advantages over a traditional static test set based approach: 
\begin{itemize}
    \item It does not require production data to be labeled in order to get an indication of production performance of the model. 
    \item It is less fragile, thus remains a better indicator of accuracy as production data changes over time
    \item It enables assessing risk of deploying a model into different deployment zones or usage scenarios, such as deploying a model into different geographic regions
\end{itemize}



\subsection{Deploy}

The goal of the deployment stage is to enable a seamless rollout of new models, with as little risk as possible.  Best practices in the continuous delivery of software services is to use {\it safe deployment} techniques, such as A/B tests, and canary releases~\cite{Istio}.
Various techniques can be used to automate the rollback decision process, significantly reducing the risk of deploying a new model.  Multi-armed bandits are one such example~\cite{Seldon}.


However, these techniques require a meaningful signal of quality in order to detect problems and trigger rollback reliably, and such signals are often not readily available in a typical model deployment.
Our AI enabled deploy stage leverages both performance prediction and KPI analytics as the metrics for driving safe deployment. These metrics provide reliable and automated indicators of model quality with respect to the application, and therefore are both effective and efficient for identifying potential problems and triggering rollback in production.

\subsection{Monitor}
The main objective during the monitoring stage is to manage the risks of in-production models by checking for performance drift~\cite{Moreno2012,Rodriguez2008}
and alerting an operator that model accuracy has dropped. 
Traditionally, this is achieved by performing periodic manual assessments of the log data. This is labor intensive, and thus expensive and tedious to perform on a regular basis.   {\it Feature drift} (also referred to as {\it covariate shift}) ~\cite{47967, cortes2008sample} and {\it prior shift} \cite{card2018}, are relatively straightforward to detect, neither of these necessarily implies that model accuracy has decreased and can produce many false positives~\cite{kim2018guiding, 1565737, zhang2019noisesensitivityanalysisbased}.

Our AI enabled monitoring approach is focused on detecting drift that {\it actually impacts model performance}.
It does so by leveraging performance prediction and KPI analytics.  Performance prediction is run periodically
on production traffic. If the predicted accuracy of the model drops, or if a KPI metric drops, an alert is triggered pulling in human assistance to perform an analysis of the relevant models.

\subsection{Diagnose and Improve}
Model improvement is typically an iterative, continuous process,
and is known to be a challenging task that requires skill and expertise.
There are many ways of improving AI models, such as feature engineering, model architecture selection, hyperparameter tuning, and the addition of more training data using techniques such as active learning ~\cite{settles2009active}.

Our diagnosis and improve stage is driven by a KPI analysis, which strives to  reduce the time and expertise needed to identify root causes and improve performance.
First, KPI analysis can be used to narrow down the scope of the problem, by focusing diagnosis efforts on data points that were involved in low-KPI transactions. This reduces the volume of data that needs to be examined and focuses the human analysis on the problematic cases.

But even more importantly, KPI analysis allows comparing and contrasting model metrics across the good versus bad transactions.  Simple correlation analysis can identify model trends that correlate with KPI drops, and thus are worthy of human investigation.  More advanced causal analyses may also be possible, depending on the available  information about the relevant models~\cite{schlkopf2019causality}.

\section{Conclusions}

Today's AI operations pipelines require significant human intervention and effort.   This paper described a set of enabling technologies that help increase the level of automation during AI operations, thus reducing the human effort and cost required.  We showed how these technologies can be used to drive automation within four common operations pipeline stages - pre-deploy test, deploy, monitoring, and improvement.
We believe there is significant opportunity remaining for improving AI operations automation and hope to encourage more research in this area. 

\bibliographystyle{abbrv}
\bibliography{main.bib}

\begin{thebibliography}{10}

\bibitem{Istio}
{\em Istio: Connect, secure, control, and observe services}.

\bibitem{Seldon}
{\em Seldon: Open source machine learning deployment}.

\bibitem{Rodriguez2008}
R.~Alaiz-Rodrıguez and N.~, Japkowicz.
\newblock Assessing the impact of changing environments on classifier
  performance.
\newblock In {\em Proceedings of the 21st Conference on Advances in Artificial
  Intelligence, Canadian AI ’08}. Springer, 2008.

\bibitem{47967}
E.~Breck, M.~Zinkevich, N.~Polyzotis, S.~Whang, and S.~Roy.
\newblock Data validation for machine learning.
\newblock In {\em Proceedings of SysML}, 2019.

\bibitem{card2018}
D.~Card and N.~A. Smith.
\newblock The importance of calibration for estimating proportions from
  annotations.
\newblock In {\em Proceedings of the 2018 Conference of the North {A}merican
  Chapter of the Association for Computational Linguistics: Human Language
  Technologies, Volume 1 (Long Papers)}, pages 1636--1646, New Orleans,
  Louisiana, June 2018. Association for Computational Linguistics.

\bibitem{Chen2019}
T.~Chen, J.~Navr{\'{a}}til, V.~Iyengar, and K.~Shanmugam.
\newblock Confidence scoring using whitebox meta-models with linear classifier
  probes.
\newblock In {\em The 22nd International Conference on Artificial Intelligence
  and Statistics, {AISTATS} 2019, 16-18 April 2019, Naha, Okinawa, Japan},
  pages 1467--1475, 2019.

\bibitem{cortes2008sample}
C.~Cortes, M.~Mohri, M.~Riley, and A.~Rostamizadeh.
\newblock Sample selection bias correction theory.
\newblock In {\em International conference on algorithmic learning theory},
  pages 38--53. Springer, 2008.

\bibitem{dudik2005}
M.~Dudík, R.~Schapire, and S.~Phillips.
\newblock Correcting sample selection bias in maximum entropy density
  estimation.
\newblock volume~17, 01 2005.

\bibitem{gal16}
Y.~Gal and Z.~Ghahramani.
\newblock Dropout as a bayesian approximation: Representing model uncertainty
  in deep learning.
\newblock In M.~F. Balcan and K.~Q. Weinberger, editors, {\em Proceedings of
  The 33rd International Conference on Machine Learning}, volume~48 of {\em
  Proceedings of Machine Learning Research}, pages 1050--1059, New York, New
  York, USA, 20--22 Jun 2016. PMLR.

\bibitem{Geifman2017}
Y.~Geifman and R.~El-Yaniv.
\newblock Selective classification for deep neural networks.
\newblock In I.~Guyon, U.~V. Luxburg, S.~Bengio, H.~Wallach, R.~Fergus,
  S.~Vishwanathan, and R.~Garnett, editors, {\em Advances in Neural Information
  Processing Systems 30}, pages 4878--4887. Curran Associates, Inc., 2017.

\bibitem{GuoCalibration2017}
C.~Guo, G.~Pleiss, Y.~Sun, and K.~Q. Weinberger.
\newblock On calibration of modern neural networks.
\newblock In {\em Proceedings of the 34th International Conference on Machine
  Learning - Volume 70}, ICML’17, page 1321–1330. JMLR.org, 2017.

\bibitem{Ishikawa2018}
F.~Ishikawa.
\newblock Concepts in quality assessment for machine learning - from test data
  to arguments.
\newblock In {\em Proceedings of the International Conference on Conceptual
  Modeling}. Springer, 2018.

\bibitem{kim2018guiding}
J.~Kim, R.~Feldt, and S.~Yoo.
\newblock Guiding deep learning system testing using surprise adequacy, 2018.

\bibitem{Moreno2012}
J.~G. Moreno-Torres, T.~Raeder, A.-R. Rocio, N.~Chawla, and F.~Herrera.
\newblock A unifying view on dataset shift in classification.
\newblock {\em Pattern Recognition}, 45, 2012.

\bibitem{schlkopf2019causality}
B.~Schölkopf.
\newblock Causality for machine learning, 2019.

\bibitem{settles2009active}
B.~Settles.
\newblock Active learning literature survey.
\newblock Technical report, University of Wisconsin-Madison Department of
  Computer Sciences, 2009.

\bibitem{1565737}
{Wei Fan}, I.~{Davidson}, B.~{Zadrozny}, and P.~S. {Yu}.
\newblock An improved categorization of classifier's sensitivity on sample
  selection bias.
\newblock In {\em Fifth IEEE International Conference on Data Mining
  (ICDM'05)}, pages 4 pp.--, Nov 2005.

\bibitem{Zadrozny2019}
B.~Zadrozny.
\newblock Learning and evaluating classifiers under sample selection bias.
\newblock In {\em Proceedings of the Twenty-First International Conference on
  Machine Learning}, ICML ’04, page 114, New York, NY, USA, 2004. Association
  for Computing Machinery.

\bibitem{ZadroznyCalibration2002}
B.~Zadrozny and C.~Elkan.
\newblock Transforming classifier scores into accurate multiclass probability
  estimates.
\newblock In {\em Proceedings of the Eighth ACM SIGKDD International Conference
  on Knowledge Discovery and Data Mining}, KDD ’02, page 694–699, New York,
  NY, USA, 2002. Association for Computing Machinery.

\bibitem{zhang2019machine}
J.~M. Zhang, M.~Harman, L.~Ma, and Y.~Liu.
\newblock Machine learning testing: Survey, landscapes and horizons, 2019.

\bibitem{zhang2019noisesensitivityanalysisbased}
L.~Zhang, X.~Sun, Y.~Li, and Z.~Zhang.
\newblock A noise-sensitivity-analysis-based test prioritization technique for
  deep neural networks, 2019.

\end{thebibliography}
\end{document}